\documentclass[11pt]{article}

\usepackage[preprint]{acl}

\usepackage{times}
\usepackage{latexsym}
\usepackage{tcolorbox}
\usepackage{tcolorbox}
\usepackage{enumerate}
\usepackage{enumitem}
\usepackage{amsmath}
\usepackage{multirow}
\usepackage{booktabs}
\usepackage{subcaption}
\usepackage{textcomp}

\usepackage[LGR,T1]{fontenc}
\usepackage[greek,english]{babel}

\usepackage[T1]{fontenc}

\usepackage[utf8]{inputenc}

\usepackage{microtype}

\usepackage{inconsolata}

\usepackage{graphicx}

\title{Semantic Label Drift in Cross-Cultural Translation}

\author{
 \textbf{Mohsinul Kabir\textsuperscript{$\pmb\dagger$}},
 \textbf{Tasnim Ahmed\textsuperscript{$\clubsuit$}},
 \textbf{Md Mezbaur Rahman\textsuperscript{$\spadesuit$}},\\
 \textbf{Polydoros Giannouris\textsuperscript{$\pmb\dagger$}},
 \textbf{Sophia Ananiadou\textsuperscript{$\pmb\dagger$}}
\\
 \textsuperscript{$\pmb\dagger$}Department of Computer Science, National Center for Text Mining,\\ The University of Manchester\\
 \textsuperscript{$\clubsuit$}School of Computing, Queen’s University, Ontario, Canada\\
 \textsuperscript{$\spadesuit$}Computer Science, University of Illinois Chicago\\
 \texttt{\normalsize{
   \{mdmohsinul.kabir, sophia.ananiadou, polydoros.giannouris\}@manchester.ac.uk,}}
   \\  \texttt{\normalsize{tasnim.ahmed@queensu.ca, mrahma56@uic.edu}}
}

\begin{document}
\maketitle
\begin{abstract}
Machine Translation (MT) is widely employed to address resource scarcity in low-resource languages by generating synthetic data from high-resource counterparts. While sentiment preservation in translation has long been studied, a critical but underexplored factor is the role of cultural alignment between source and target languages. In this paper, we hypothesize that semantic labels are drifted or altered during MT due to cultural divergence. Through a series of experiments across culturally sensitive and neutral domains, we establish three key findings: (1) MT systems, including modern Large Language Models (LLMs), induce label drift during translation, particularly in culturally sensitive domains; (2) unlike earlier statistical MT tools, LLMs encode cultural knowledge, and leveraging this knowledge can amplify label drift; and (3) cultural similarity or dissimilarity between source and target languages is a crucial determinant of label preservation. Our findings highlight that neglecting cultural factors in MT not only undermines label fidelity but also risks misinterpretation and cultural conflict in downstream applications.\\
\textit{This paper includes examples that may contain offensive or sensitive language, presented solely for research and demonstration purposes.}
\end{abstract}

\section{Introduction}

Machine Translation (MT) has long served as a vital technology for enabling cross-cultural communication, with applications in chat systems, customer support, and social media. Beyond direct communication, MT plays a crucial role in research by facilitating dataset reuse across languages, particularly in low-resource settings. For many low-resource languages in regions such as South Asia and Africa, translating English datasets has become a common strategy to support NLP development \citep{steigerwald2022overcoming, nekoto2020participatory}. This approach allows these linguistic communities to benefit from shared knowledge and mitigate data scarcity. With the emergence of large language models (LLMs), this paradigm has become even more prevalent as translated data are now widely used for benchmarking tasks and for both pretraining and fine-tuning LLMs across diverse linguistic domains.

A critical concern in such dataset reuse is the preservation of semantic labels during translation. Prior studies \citep{memon2021impact, mohammad2016translation} have shown that translating datasets from high-resource to low-resource languages can yield high label preservation, with only $2-3\%$ degradation. However, these findings apply largely to straightforward sentiment analysis tasks with binary labels (positive/negative), where cultural divergence plays a minimal role. In more nuanced research areas, such as affective computing, subtler linguistic cues become crucial. While LLMs and modern MT systems perform well in translating factual or \textit{propositional} content, they still struggle with \textit{non-propositional} aspects, such as politeness, formality, or emotion. Studies by \citet{mirkin2015motivating} and \citet{rabinovich2016personalized} reveal that translation can obscure socio-demographic cues like gender and personality traits, while \citet{troiano2020lost} demonstrates that transformer-based NMT systems often lose emotional content during translation. Similarly, \citet{havaldar2025towards} show that cultural differences can cause misalignment between a speaker’s intended style and a listener’s interpretation; for example, politeness is often lost in translation.

Translating text into culturally sensitive domains, such as mental health discourse, sarcasm, or irony, requires more than literal fidelity. It demands the preservation of affective and stylistic elements, such as \textit{empathy} and \textit{tone}, to bridge cultural gaps effectively. Despite this, research remains limited on how translation may alter emotional polarity or even cause label drift between source and target languages, including when translations are generated by state-of-the-art (SOTA) LLMs. In this study, we explore the hypothesis that semantic label drift predominantly occurs in affective and culturally sensitive contexts, systematically examining failures in label preservation during cross-cultural translation. Our experiments include both traditional statistical MT systems and contemporary LLMs as translation tools. Specifically, we address the following four research questions: 

\begin{enumerate}[label=RQ\arabic*:, leftmargin=*]
    \item Do state-of-the-art machine translation systems (e.g., Google Translate, NLLB, and modern LLMs) alter dataset labels in culturally sensitive domains during translation? (Yes)
    
    \item Does the cultural knowledge of LLMs reduce or amplify label drift between source and target languages? (May amplify)
    
    \item Does cultural similarity between source and target languages mitigate label drift? (Yes, but in specific domains)

    \item Is label drift in translated texts culture-specific, or does it also occur in culturally neutral domains? (Mostly culture-specific)
\end{enumerate}

In addition to analyzing label preservation quantitatively, we also perform a qualitative analysis of translated texts, identifying critical cases of translation refusal by LLMs and discussing instances of cultural misalignment that may result in culturally inappropriate or distorted translations.

\section{Related Works}

\textbf{Statistical Machine Translation and Sentiment Alteration.} 
Statistical Machine Translation (SMT) systems such as Google Translate remain widely used and effective for high-resource languages. However, true human-like translation requires a deeper understanding of semantics and context \citep{weaver1952translation}, which SMT and even Neural Machine Translation (NMT) models often fail to capture, particularly regarding emotional nuance and cultural subtleties. Several studies have shown that MT can alter sentiment polarity. For example, \citet{salameh2015sentiment} report that Arabic social media posts translated into English frequently lose or neutralize their original sentiment. Similarly, \citet{saadany2023analysing} find that online NMT systems often invert or erase emotional tone in tweets. To address this, \citet{kumari2021reinforced} propose fine-tuning global-attention NMT models using actor–critic reinforcement learning with sentiment- and semantics-based rewards, improving preservation in low-resource translation.

\textbf{Cultural Codes of Language.} 
Although basic human behaviors are universal, their linguistic expressions are deeply culture-specific. Each language encodes norms, prohibitions, and imperatives differently. From this perspective, \citet{aydarova2024axiological} analyze the axiological (value-related) and cultural codes embedded in behavioral verbs across Russian, Tatar, and English, revealing both shared and unique semantic patterns. They argue that overlooking such codes can lead to cultural misinterpretation in MT. \citet{troiano2020lost} empirically demonstrate that MT systems systematically degrade non-propositional emotional information and propose a re-ranking method to mitigate this loss. More recently, \citet{havaldar2025towards} show that effective cross-cultural translation is essential to convey affective styles such as politeness or intimacy, as these are shaped by cultural norms. Their findings indicate that even modern LLMs frequently fail to preserve such stylistic cues, often neutralizing or misrendering them, particularly in non-Western languages.

\section{Evaluating Label Preservation in Cross-Cultural MT}

\subsection{Culturally Sensitive Domains}

We hypothesize that label alteration occurs primarily in culturally sensitive domains. To select appropriate domains, we consider two factors: cultural sensitivity established in prior literature and availability of datasets. Based on these, we focus on mental health and irony. Mental health concepts, diagnoses, and treatments are shaped by cultural beliefs and practices \citep{rai2024cross, adebayo2024cross}, with numerous studies reporting stigma across individualistic and collectivist societies and the risks of cultural insensitivity in diagnosis, assessment, or intervention \citep{kirmayer1989cultural, altweck2015mental, lyons2025culturally}. Likewise, irony is present across languages but varies in definition, function, and recognition across cultures \citep{weizman2022explicitating}, shifting from explicit to subtle forms between individualistic and collectivist contexts \citep{ervas2024irony}.

To analyze label shift, we require granular datasets in these two domains with multi-class annotations, primarily based on data collected from \textit{western} users. Accordingly, we select the following datasets:

\begin{enumerate}
    \item \textbf{DEPTWEET} \citep{kabir2023deptweet}: tweets labeled into four depressive categories (\textit{Non-depressed}, \textit{Mildly depressed}, \textit{Moderately depressed}, \textit{Severely depressed}).

    \item \textbf{SemEval-2018 Task 3} \citep{van2017can}: tweets labeled into four irony categories (\textit{Non-ironic}, \textit{Irony by clash}, \textit{Situational irony}, \textit{Other irony}).  

\end{enumerate}

We randomly select $1150$ samples from each dataset for our experiments, ensuring an equal number of instances from each class. However, the \textit{Situational irony} and \textit{Other irony} classes in the original \textbf{SemEval-2018 Task 3} dataset contain only about $200$ samples, so we include all available instances from these classes.

\begin{figure*}[ht]
    \centering
    \includegraphics[width=\textwidth]{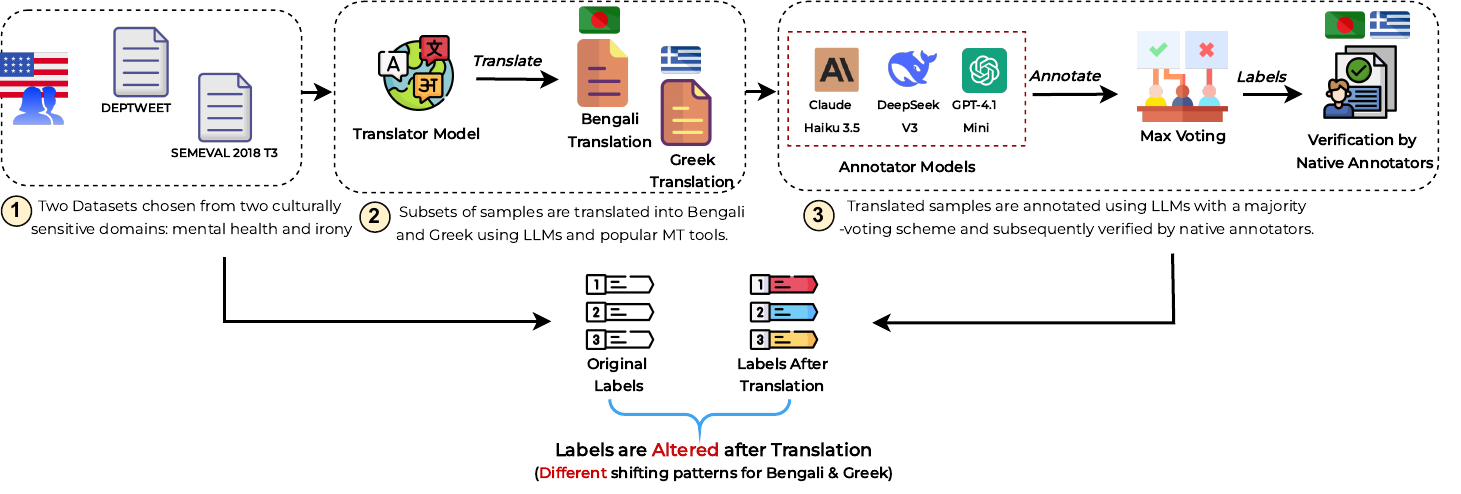}
    \caption{Evaluating label shifts in cross-cultural translation. We select two culturally sensitive datasets constructed in Western contexts, translate subsets of randomly sampled data into Bengali and Greek, and annotate them with verification by native speakers. By comparing translated annotations with the original labels, we assess the extent of label preservation during translation.}
    \label{fig:label_shift_full}
\end{figure*}

\subsection{Models \& Probing Methods} \label{sec:models_prompting}

We aim to simulate the experience of an average user relying on online translation tools. To this end, we evaluate a mix of translation systems alongside both open-source and proprietary LLMs. Specifically, our experiments include Google Translate, NLLB-1.3B, GPT-4.1 Mini, Claude Sonnet 4, Llama-3.3 (70B), and DeepSeek V3.

\begin{figure}[ht]
\small
\centering
\begin{tcolorbox}[colback=blue!5,colframe=blue!40!black, title=]
Imagine you are a
\{marital\_status\} \{sex\} from
\{region\}, \{country\}.
You are \{age\} years of age
and completed \{education\}
education level.
You consider yourself part
of the \{social\_class\}.\\
Translate the following English sentence into \{target\_language\} from this
perspective.\\
\{Sentence:\}

\end{tcolorbox}
\caption{Translation with Anthropological Prompting. }    
\label{fig:anthro_prompt}
\end{figure}

For LLMs, we design two prompting strategies. In the first, we mimic typical user behavior by issuing a simple instruction: \textit{"Provide translation of the following sentence from English to \{target\_language\}."} Translations generated this way are referred to as \textit{Literal Translation}.

In the second, we leverage models’ cultural knowledge through \textit{Anthropological Prompting} \citep{alkhamissi2024investigating}, which grounds instructions in anthropological contexts by asking the model to assume a culturally situated perspective. The method uses six demographic dimensions, with empirical evidence showing optimal cultural alignment under the following configuration: \texttt{Region}: Country-Specific, \texttt{Sex}: Male, \texttt{Age}: $<50$, \texttt{Social Class}: Upper/Lower Middle Class, \texttt{Education Level}: Higher, and \texttt{Marital Status}: Married. We adopt this setup to maximize cultural alignment, and translations generated in this way are referred to as \textit{Cultural Prompting} (see Figure \ref{fig:anthro_prompt}).

\subsection{Country \& Language Selection}

Although using language as a direct proxy for culture has been criticized \citep{kabir2025break}, it remains unavoidable in the context of translation tasks. We hypothesize that when the source and target languages belong to culturally similar contexts, dataset labels are better preserved, with fewer alterations. Since both datasets in our study originate primarily from North-American and European user data, we select one culturally similar language and one contrasting language. For cultural comparison, we rely on the Inglehart-Welzel World Cultural Map \citep{WVSInglehartWelzel2023}, which is structured around two dimensions: Traditional vs. Secular-rational values and Survival vs. Self-expression values. We also prioritize low-resource languages, as translation is especially relevant for such domains. Based on these criteria, we choose \textbf{Greek}, representing an individualistic culture similar to the Anglo-centric world, and \textbf{Bengali}, representing a contrasting collectivist culture.

\begin{table*}[ht]
\centering
\scriptsize
\setlength{\tabcolsep}{5pt}
\renewcommand{\arraystretch}{1.2}
\resizebox{\textwidth}{!}{%

\begin{tabular}{lccccccccc}
\specialrule{1.5pt}{0pt}{0pt}
\multicolumn{10}{c}{\textbf{DEPTWEET Dataset (Mental Health)}} \\
\midrule
\multicolumn{10}{c}{\textbf{Bengali Translation}} \\
\midrule
& \multicolumn{4}{c}{\textbf{\textit{Literal Prompting}}} & \multicolumn{4}{c}{\textbf{\textit{Cultural Prompting}}} & \multirow{2}{*}{\textbf{$\Delta$(Lit.-Cul.) }} \\
\cmidrule(lr){2-5} \cmidrule(lr){6-9}
& \textbf{Non-Depressed} & \textbf{Mild} & \textbf{Moderate} & \textbf{Severe} & \textbf{Non-Depressed} & \textbf{Mild} & \textbf{Moderate} & \textbf{Severe} & \\
\midrule
Google Translate & 44.3 & 50.7 & 58.3 & 84.7 & - & - & - & - & - \\
NLLB-1.3B & \textbf{56.4} & 53.1 & 54.9 & 84.3 & - & - & - & - & - \\
Claude-4-Sonnet & 47.7 & 46.9 & \textbf{62.2} & 82.6 & 48.4 & 48.6 & \textbf{63.5} & 76.0 & +0.73  \\
GPT-4.1-Mini & 43.9 & 52.4 & 61.1 & 86.1 & 44.6 & 56.2 & 60.1 & 78.7 & +0.98  \\
Llama-3.3 & 48.4 & 49.0 & 53.5 & \textbf{87.8} & 48.1 & 48.6 & 53.5 & \textbf{87.5} & +0.25  \\
DeepSeek V3 & 54.0 & \textbf{58.7} & 55.9 & 84.3 & \textbf{54.7} & \textbf{58.7} & 58.0 & 82.2 & -0.18  \\
\midrule
\multicolumn{10}{c}{\textbf{Greek Translation}} \\
\midrule
& \multicolumn{4}{c}{\textbf{\textit{Literal Prompting}}} & \multicolumn{4}{c}{\textbf{\textit{Cultural Prompting}}} & \multirow{2}{*}{\textbf{$\Delta$(Lit.-Cul.) }}\\
\cmidrule(lr){2-5} \cmidrule(lr){6-9}
& \textbf{Non-Depressed} & \textbf{Mild} & \textbf{Moderate} & \textbf{Severe} & \textbf{Non-Depressed} & \textbf{Mild} & \textbf{Moderate} & \textbf{Severe} & \\
\midrule
Google Translate & 40.4 & 51.4 & 57.6 & 84.0 & - & - & - & - & - \\
NLLB-1.3B & \textbf{57.1} & \textbf{56.9} & 55.6 & 77.7 & - & - & - & - & - \\
Claude-4-Sonnet & 48.1 & 48.3 & 61.8 & 86.1 & 49.5 & 49.0 & \textbf{62.2} & 81.9 & +0.43  \\
GPT-4.1-Mini & 48.4 & 50.3 & \textbf{62.2} & 86.8 & 48.8 & 52.1 & 60.8 & 79.1 & +1.73 \\
Llama-3.3 & 42.2 & 47.9 & 54.9 & \textbf{88.9} & 42.9 & 46.2 & 52.1 & \textbf{85.4} & +1.83  \\
DeepSeek V3 & 54.0 & 52.8 & 60.1 & 85.7 & \textbf{53.7} & \textbf{52.4} & 60.4 & 85.4 & +0.18 \\
\midrule
\textit{$\Delta$ \textbf{(Greek-Bengali)$_{deptweet}$}} & -0.75 & -0.53 & +1.05 & -0.10 & -0.23 & -3.10 & +0.10 & +1.85 & \\
\specialrule{1.5pt}{0pt}{0pt}
\midrule
\multicolumn{10}{c}{\textbf{SemEval-2018 T3 Dataset (Irony)}} \\
\midrule
\multicolumn{10}{c}{\textbf{Bengali Translation}} \\
\midrule
& \multicolumn{4}{c}{\textbf{\textit{Literal Prompting}}} & \multicolumn{4}{c}{\textbf{\textit{Cultural Prompting}}} & \multirow{2}{*}{\textbf{$\Delta$(Lit.-Cul.) }} \\
\cmidrule(lr){2-5} \cmidrule(lr){6-9}
& \textbf{Non-ironic} & \textbf{Ironic-clash} & \textbf{Situational-irony} & \textbf{Other-irony} & \textbf{Non-ironic} & \textbf{Ironic-clash} & \textbf{Situational-irony} & \textbf{Other-irony} & \\
\midrule
Google Translate & 29.7 & \textbf{81.8} & 21.0 & 0.5 & - & - & - & - & - \\
NLLB-1.3B & 53.1 & 56.7 & \textbf{31.3} & \textbf{3.0} & - & - & - & - & - \\
Claude-4-Sonnet & 48.9 & 65.3 & 25.2 & 2.5 & 53.4 & \textbf{64.7} & 19.2 & 0.0 & +1.15 \\
GPT-4.1-Mini & 51.4 & 67.1 & 22.0 & 2.5 & 54.4 & 62.6 & 22.0 & \textbf{3.0} & +0.25  \\
Llama-3.3 & \textbf{62.2} & 47.2 & 20.6 & 2.0 & 55.4 & 54.6 & 21.0 & 2.5 & -0.38 \\
DeepSeek V3 & 58.4 & 62.3 & 28.5 & 1.0 & \textbf{58.4} & 62.3 & \textbf{28.0} & 1.0 & +0.13  \\
\midrule
\multicolumn{10}{c}{\textbf{Greek Translation}} \\
\midrule
& \multicolumn{4}{c}{\textbf{\textit{Literal Prompting}}} & \multicolumn{4}{c}{\textbf{\textit{Cultural Prompting}}} & \multirow{2}{*}{\textbf{$\Delta$(Lit.-Cul.) }}\\
\cmidrule(lr){2-5} \cmidrule(lr){6-9}
& \textbf{Non-ironic} & \textbf{Ironic-clash} & \textbf{Situational-irony} & \textbf{Other-irony} & \textbf{Non-ironic} & \textbf{Ironic-clash} & \textbf{Situational-irony} & \textbf{Other-irony} & \\
\midrule
Google Translate & 32.1 & \textbf{78.3} & 21.5 & 2.5 & - & - & - & - & - \\
NLLB-1.3B & 51.6 & 59.3 & \textbf{33.2} & 3.5 & - & - & - & - & - \\
Claude-4-Sonnet & 58.4 & 69.4 & 28.0 & 3.0 & \textbf{62.4} & 64.4 & 25.7 & 4.0 & +0.58 \\
GPT-4.1-Mini & 59.1 & 71.2 & 29.9 & 3.5 & 61.4 & 67.7 & \textbf{29.9} & 3.5 & +0.30 \\
Llama-3.3 & \textbf{64.2} & 57.6 & 21.0 & \textbf{4.0} & 55.4 & 60.2 & 18.7 & 3.0 & +2.38  \\
DeepSeek V3 & 57.1 & 71.8 & 25.7 & 4.0 & 56.1 & \textbf{73.0} & 24.8 & \textbf{4.0} & +0.18 \\
\midrule
\textit{$\Delta$ \textbf{(Greek-Bengali)$_{semeval2018}$}} & +3.13 & +4.53* & +1.78 & +1.50* & +3.43 & +5.28 & +2.23 & +2.00 & \\
\specialrule{1.5pt}{0pt}{0pt}
\end{tabular}
}
\caption{Label preservation rates (\%) for the \textbf{Mental Health} (top) and \textbf{Irony} (bottom) datasets translated into Bengali and Greek using six MT models. Results are shown for both \textit{literal} and \textit{cultural} prompting. \textbf{Bold} numbers indicate the highest preservation per class. Mean differences between Greek and Bengali ($\Delta$(Greek–Bengali)) and between prompting types ($\Delta$ (Lit.-Cul.) are reported. Statistical significance is assessed via one-sided Wilcoxon signed-rank tests ($^{*}p<0.05$). No significant gain is observed for cultural over literal prompting; only \textit{Ironic-clash} and \textit{Other-irony} show significant cross-lingual differences.}

\label{tab:label_drift}
\end{table*}


\subsection{Translation \& Annotation} \label{sec:trans_and_annot}

We adopt a Human–LLM collaboration scheme \citep{wang2024human} to annotate translated samples. For consistency, we follow the annotation schemes described in the original DEPTWEET \citep{kabir2023deptweet} and SemEval-2018 T3 \citep{van2017can} papers, where authors detail their annotation procedures. These schemes are converted into detailed annotation guidelines, which we then use as prompting instructions for LLM-based annotation. LLMs have been shown to produce high-quality annotations in diverse NLP tasks, often rivaling or surpassing crowdsourced annotators \citep{He2023AnnoLLM, Tan2024Large}. However, given the complexity of mental health and irony, which require nuanced interpretation, we adopt the following strategy:

\begin{enumerate}
\item \textbf{Majority Voting:} Each sample is annotated by three LLMs: GPT-4.1 Mini, Claude Haiku 3.5, and DeepSeek V3, selected for their cost and speed-efficiency. The final label is determined via majority voting.
\item \textbf{Human Validation:} Native speakers of Bengali and Greek review ambiguous cases (e.g., instances without consensus such as three distinct labels). They also annotate a subset of translated samples to validate the LLM-derived majority-vote labels.

\end{enumerate}

This Human–LLM collaboration approach balances cost-effectiveness with reliability in complex domains that is best suited for our study. \citep{wang2024human}. Our full experiment pipeline is demonstrated in Figure \ref{fig:label_shift_full}.

\subsection{Evaluation Metrics}

To assess the consistency between the original and translated labels, we employ three complementary evaluation metrics:

\begin{itemize}
    \item \textbf{Label Preservation Rate}: Measures the proportion of labels that remain unchanged after translation.
    \item \textbf{Kullback–Leibler (KL) Divergence}: Captures the overall distributional shift of labels between the source and translated datasets.
    \item \textbf{Matthews Correlation Coefficient (MCC)}: Evaluates the strength of correspondence between original and translated labels.
\end{itemize}

\begin{figure*}
    \centering
    \includegraphics[width=\textwidth]{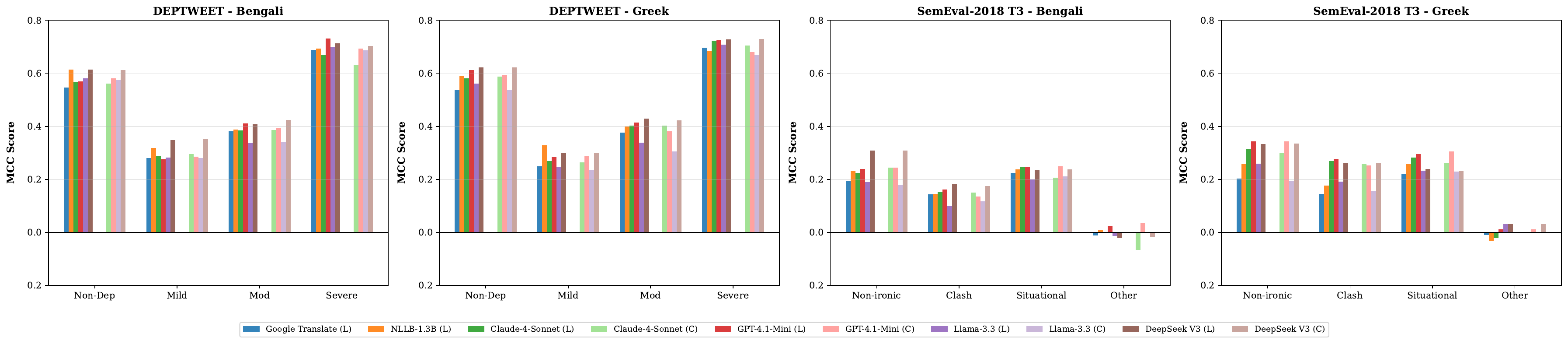}
    \caption{Matthews Correlation Coefficient (MCC) scores comparing the performance of six translation/language models. The grouped bar charts display literal prompting (L) results (darker bars) for all models, while cultural prompting (C) results (lighter bars) are shown only for the four LLM-based models. MCC scores interpret as: $0.0-0.3:$weak, $0.31-0.5:$ weak moderate, $0.51-0.7:$ moderate strong, and $>0.7:$ strong performance. }
    \label{fig:mcc_all}
\end{figure*}

\section{Findings}

\subsection*{RQ1. Translation Preserves Perceived Severity but Obscures Mild Presentations}

Table \ref{tab:label_drift} presents the label preservation rates following translation for both datasets. A clear pattern emerges where labels for the extreme classes are consistently well-preserved. For the \textbf{Mental Health} dataset, the \textit{severe} depression category exhibits high preservation rates in both Bengali and Greek. However, the translation process introduces a systematic bias towards inflating perceived severity, evidenced by the considerably lower preservation rates for the \textit{mild} and \textit{moderate} classes. This indicates that \textit{non-severe} cases are frequently drifted to more \textit{severe} categories after translation.

This finding is further quantified by calculating the Matthews Correlation Coefficient (MCC) to measure the agreement between original and post-translation labels. As shown in Figure \ref{fig:mcc_all}, strong agreement is observed for the \textit{Non-Depressed} and \textit{Severe} classes. In contrast, the \textit{Mild} and \textit{Moderate} classes show weak agreement, with much lower MCC scores. This confirms that the nuanced linguistic distinctions between mild and moderate depression are largely lost or altered during translation, making reliable classification of these categories after translation exceedingly difficult.

A similar but distinct pattern is observed for the \textbf{Irony} dataset. The \textit{non-ironic} and \textit{ironic-clash} categories are well-preserved across languages, suggesting that translation robustly preserves non-irony and obvious, clash-based irony. Conversely, the more context-dependent \textit{situational irony} is highly degraded. The MCC scores tell a more comprehensive story, revealing weak to moderate agreement for all irony categories, indicating that ironic content is generally not robust to machine translation. Most notably, the \textit{other-irony} category frequently yields MCC scores near or below zero, demonstrating that its classification after translation becomes effectively unpredictable and uncorrelated with the original labels.

\begin{figure}
    \centering
    \includegraphics[width=\columnwidth]{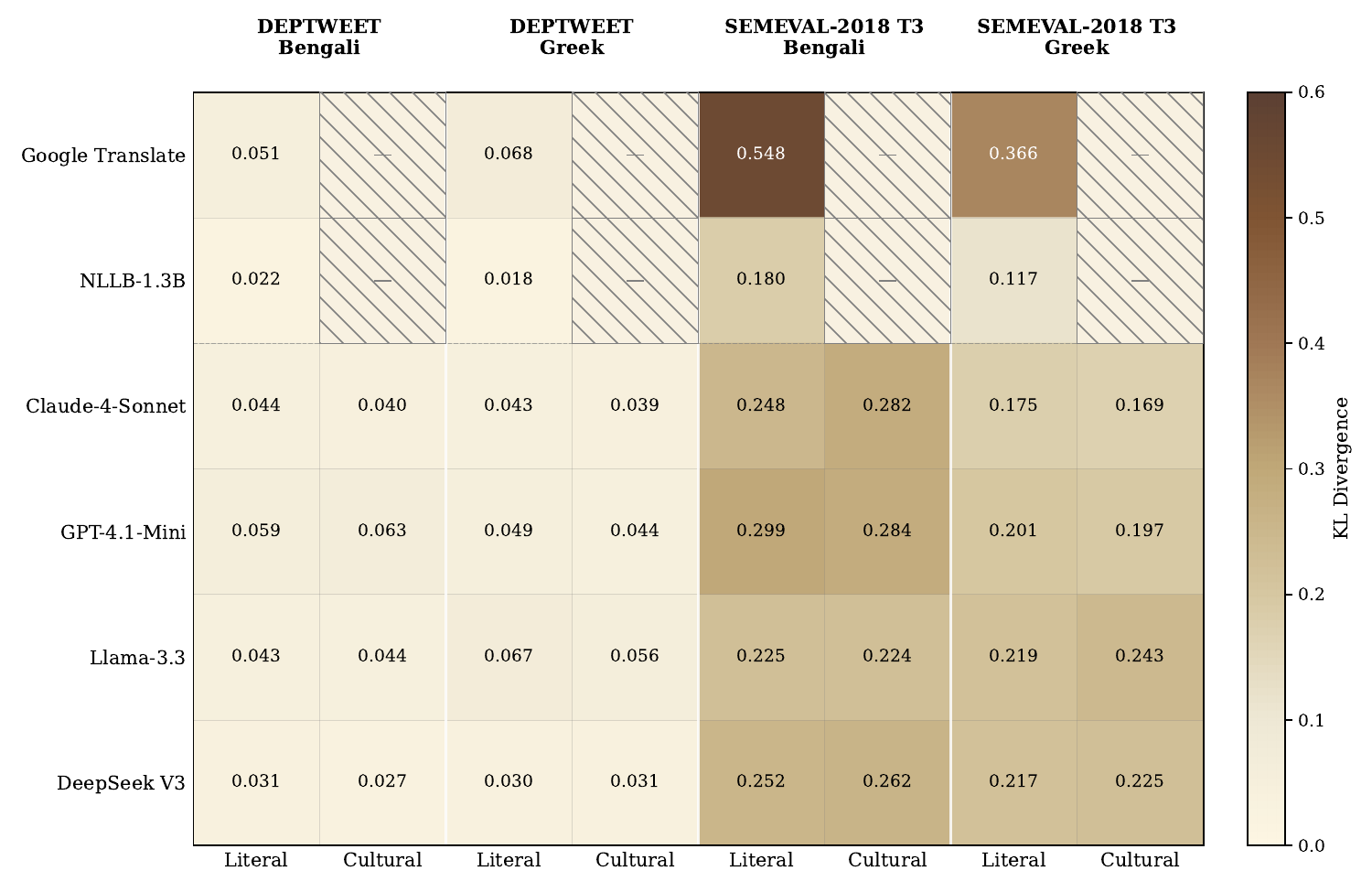}
   \caption{Kullback–Leibler (KL) divergence between original and translated label distributions across six MT models under different prompting techniques. The SemEval-2018 T3 dataset exhibits greater divergence from the original labels than DEPTWEET.}
    \label{fig:kl_div}
\end{figure}

To understand the macro-level effect of translation on the entire dataset, we report the Kullback–Leibler (KL) Divergence between the original and translated label distributions in Figure \ref{fig:kl_div}. For the \textbf{Mental Health dataset}, we observe low KL Divergence scores (all < $0.08$). This indicates that while individual instances are altered (as shown by low preservation and MCC for mild/moderate classes), the global distribution of severity labels remains largely intact. This suggests a systematic re-categorization within the severity spectrum rather than a wholesale shift. In stark contrast, the \textbf{Irony} dataset exhibits very high KL Divergence (up to $0.65$). This confirms that translation induces a severe distortion of the entire label distribution, fundamentally altering the proportion of \textit{ironic} to \textit{non-ironic} content and between different irony types. This macro-level distortion complements our instance-level findings, illustrating that the translation process has a catastrophic effect on the fabric of ironic discourse.

\subsection*{RQ2. Cultural Knowledge in LLMs may Exacerbate Label Shift during Translation}

We ground the LLM responses in relative cultural context by using the \textit{anthropological} prompting strategy shown in Figure \ref{fig:anthro_prompt} to answer RQ2. Initial results in Table \ref{tab:label_drift} show a slightly higher label preservation rate for literal prompting compared to cultural prompting, though this difference is not statistically significant.

\begin{table}[htbp]
\centering
\resizebox{\columnwidth}{!}{
\begin{tabular}{lccc}
\specialrule{1.5pt}{0pt}{0pt}
\textbf{Dataset} & \textbf{$\Delta$MCC (Lit.-Cul.)} & \textbf{p-value (wilcoxon)} \\ 
\hline
DT-Bengali & 0.0044 & 0.2589\\
DT-Greek & 0.0142$^*$ & 0.0035\\
SE T3-Bengali & 0.0041 & 0.5921\\
SE T3-Greek & 0.0112$^*$ & 0.0207\\
Overall & 0.0085$^*$ & 0.0026\\
\specialrule{1.5pt}{0pt}{0pt}
\end{tabular}
}
\caption{Mean MCC differences (Literal - Cultural) for DEPTWEET (DT) and SemEval-2018 T3 (SE T3) datasets. Positive values indicate superior performance of literal prompting. Statistical significance is assessed with a Wilcoxon signed-rank test (* $p < 0.05$).}
\label{tab:mcc_lit_cul_diff}
\end{table}

To quantify this effect, we perform a paired t-test on the Matthew's Correlation Coefficient (MCC) scores to compare the correlation of literal and cultural prompting with the original labels. A Wilcoxon signed-rank test (Table \ref{tab:mcc_lit_cul_diff}) reveals a small but statistically significant overall advantage for literal prompting over cultural prompting (Mean $\Delta\text{MCC}_{\text{Literal - Cultural}} = +0.0085$, 95\% CI [0.0039, 0.0132], $p < 0.01$).

This overall effect is primarily driven by Greek translations, where literal prompting demonstrates significantly stronger agreement with original labels on both mental health ($\Delta\text{MCC} = +0.0142$, $p < 0.01$) and irony ($\Delta\text{MCC} = +0.0112$, $p = 0.02$) tasks. In contrast, for Bengali translations, we find no significant difference between prompting strategies for either task (both $p > 0.13$).

Based on these observations, we conclude that explicitly instructing models to consider cultural context, while intended to improve relevance, can instead motivate label shift by altering the textual features critical for classification. The superior performance of literal prompting suggests that for the specific task of preserving pre-defined labels, a direct translation approach is unexpectedly more reliable and robust.

\subsection*{RQ3:  Cultural Similarity between Source and Target Languages can Mitigate Label
Shift in Specific Domains}

To investigate the role of cultural similarity, we select language pairs with varying proximity to Anglo-American culture: Greek (a European language with strong historical ties) and Bengali (a South Asian language with a distinct cultural context). We hypothesize that translations into culturally closer languages would experience less label shift.

\begin{table}[ht]
\centering
\tiny
\resizebox{\columnwidth}{!}{
\begin{tabular}{@{}lcc@{}}
\toprule
& \multicolumn{2}{c}{\textbf{$\Delta$MCC(Greek - Bengali)}} \\
\cmidrule(lr){2-3}
\textbf{} & \textbf{Literal } & \textbf{Cultural } \\
\midrule
\textbf{DEPTWEET} & & \\
Non-Depressed & 0.002 & 0.002 \\
Mild & -0.019 & -0.032 \\
Moderate & 0.009 & -0.008 \\
Severe & 0.012 & 0.017 \\
Average & 0.001 & -0.005 \\
\midrule
\textbf{SemEval-2018 T3} & & \\
Non-Ironic & 0.055* & 0.050 \\
Ironic Clash & 0.074* & 0.087 \\
Situational & 0.023* & 0.030 \\
Other & 0.003 & 0.025 \\
Average & 0.039* & 0.048 \\
\bottomrule
\end{tabular}
}
\caption{Mean MCC differences for Greek - Bengali across dataset categories. Positive values indicate better correlation of Greek translation with original labels. Statistical significance assessed via one-tailed Wilcoxon signed-rank test (* $p < 0.05$).}
\label{tab:mcc_greek_vs_bengali}
\end{table}

Our quantitative results provide strong but domain-specific support for this hypothesis. Table \ref{tab:label_drift} reports the mean difference in label preservation between Greek and Bengali translations for the mental health (DEPTWEET) and irony (SemEval-2018 T3) datasets. No statistically significant differences emerge in the mental health dataset, whereas the irony dataset shows significantly better preservation in the \textit{Ironic-clash} and \textit{Other-irony} categories. A consistent pattern appears in Table \ref{tab:mcc_greek_vs_bengali}, which presents differences in Matthew’s Correlation Coefficient (MCC), where positive values indicate stronger label correlation for Greek than Bengali. In the mental health domain, MCC differences are negligible under both literal ($\Delta=+0.001$) and cultural ($\Delta=-0.005$) prompting, with no significant effects at the category level. By contrast, in the irony domain, Greek translations consistently achieve higher label correlation. Under literal prompting, the average MCC is statitically higher ($\Delta$ = $+0.039$, $p<0.05$), with significant gains in the \textit{Non-Ironic}, \textit{Ironic-Clash}, and \textit{Situational-Irony} categories. Cultural prompting shows a similar, though not statistically significant, trend ($\Delta=+0.048$).

This divergence in outcomes between irony and mental health motivates a qualitative analysis to uncover underlying mechanisms. In irony, the relative success of literal prompting in Greek likely reflects cultural proximity to Anglo-American humor, where ironic intent often survives direct translation. Conversely, in the mental health domain, native speaker feedback reveals that cultural prompting can amplify emotional tone (e.g., translating “I feel uneasy” as “this is driving me crazy”) or censor explicit references to self-harm (e.g., replacing “suicide” with “dark thoughts”). These adaptations either intensify or neutralize emotional severity, leading to label drift. Overall, while cultural prompting enhances cultural naturalness through idiomatic expression, it tends to increase semantic divergence, whereas literal prompting maintains closer alignment with the source labels.

In conclusion, cultural similarity between source and target languages can reduce label shift, but its effect is domain-dependent. For conceptually aligned domains like irony, direct translation into a culturally proximal language such as Greek may be highly effective in label preservation. However, for culturally sensitive domains like mental health, cultural adaptation may introduce an additional layer of shift, offsetting the potential benefits of cultural proximity.

\subsection*{RQ4: Label Drift Predominantly Occurs in Culturally Sensitive Domains}

We investigate whether the phenomenon of label shift extends across domains regardless of cultural sensitivity. Prior studies have reported that translations of relatively simple domains, such as positive/negative sentiment, can achieve competitive accuracy with minimal label alteration \citep{mohammad2016translation, memon2021impact}. To build on this, we select a domain that is expected to involve little to no cultural sensitivity, i.e., culturally agnostic emotions. For dataset selection, we consider two criteria: (i) the dataset should contain granular labels, and (ii) the samples should be sufficiently long to make annotation appropriately challenging and complex. Based on these criteria, we choose the Amazon Product Review dataset \citep{hou2024bridging}, which consists of lengthy real-world product reviews accompanied by ratings from $1$ to $5$. To facilitate classification, we group the lowest scores ($1$ and $2$) as \textit{Negative (Neg.)}, the middle score ($3$) as \textit{Neutral (Neu.)}, and the highest scores ($4$ and $5$) as \textit{Positive (Pos.)}, following prior work by \citet{kabir2023banglabook, wang2020review}, among others. From this dataset, we randomly sample $200$ reviews per category and apply the translation and annotation procedure described in Section \ref{sec:trans_and_annot}. The average sentence length for the selected subset is $88.82$ words, making the annotation process considerably challenging for the annotator models.

\begin{table}[ht]
\centering
\resizebox{\columnwidth}{!}{
\begin{tabular}{llcccccc}
\specialrule{1.5pt}{0pt}{0pt}
 & & \multicolumn{3}{c}{\textit{\textbf{Literal Prompting}}} & \multicolumn{3}{c}{\textit{\textbf{Cultural Prompting}}} \\ 
\cline{3-8}
\textbf{Language} & \textbf{Model} & \textbf{Neg.} & \textbf{Neu.} & \textbf{Pos.} & \textbf{Neg.} & \textbf{Neu.} & \textbf{Pos.} \\
\hline
\multirow{4}{*}{\textbf{Bengali}} 
 & Google Translate & 0.910 & 0.460 & 0.895 &  --   &  --   &  --   \\
 & NLLB-1.3B & 0.905 & 0.460 & 0.910 &  --   &  --   &  --   \\
 & Claude-4-Sonnet  & 0.905 & 0.475 & 0.900  & 0.905 & \textbf{0.480}  & 0.905 \\
 & GPT-4.1-Mini     & 0.915 & \textbf{0.480}  & 0.905  & 0.910 & 0.475 & 0.915 \\
 & Llama-3.3 & 0.910 & 0.470 & \textbf{0.915} & \textbf{0.920} & 0.465 & \textbf{0.920} \\
 & DeepSeek V3      & \textbf{0.915} & 0.445 & 0.910  & 0.905 & 0.445 & 0.910 \\
\toprule
\multirow{4}{*}{\textbf{Greek}} 
 & Google Translate & 0.935 & 0.395 & \textbf{0.920}  &  --   &  --   &  --   \\
  & NLLB-1.3B & 0.940 & 0.465 & 0.910  &  --   &  --   &  --   \\
 & Claude-4-Sonnet  & 0.945 & \textbf{0.470}  & 0.905 & 0.940  & \textbf{0.455} & 0.900   \\
 & GPT-4.1-Mini     & 0.950  & 0.440  & 0.915 & 0.950  & 0.435 & \textbf{0.915} \\
 & Llama-3.3 & 0.930 & 0.445 & 0.910 & 0.935 & 0.435 & 0.905 \\
 & DeepSeek V3      & \textbf{0.960}  & 0.450  & 0.900   & \textbf{0.950}  & 0.450  & 0.910 \\
\specialrule{1.5pt}{0pt}{0pt}
\end{tabular}
}
\caption{Label preservation rates (\%) for Amazon Product Review dataset after translation into Bengali and Greek. Bold values indicate the highest preservation per category across models. Label preservation is notably higher than in DEPTWEET and SEMEVAL-2018 T3.}
\label{tab:amazon_results}
\end{table}

Despite these challenges, the results in Table \ref{tab:amazon_results} show very high label preservation for the \textit{Negative} and \textit{Positive} samples. The relatively lower preservation rate in the \textit{Neutral} category reflects the well-documented phenomenon of \textit{polarity shift}, where neutral emotions are translated into positive or negative emotions (and vice versa) \citep{lohar2017maintaining, hartung2023measuring}. A possible explanation is that models often struggle to assign a neutral label due to shifts in linguistic features introduced during translation \citep{salameh2015sentiment}. Our calculated Kullback-Leibler (KL) Divergence between original and translated labels ranges from $0.046$ to $0.053$ for Bengali and from $0.061$ to $0.072$ for Greek, indicating a minimal distributional shift after translation. These findings suggest that label drift is less pronounced in culturally neutral domains, where polarity effects dominate rather than cultural misalignment.

\section{Annotation Framework and Agreement Analysis}



We employ a majority-voting scheme across three LLMs for automatic annotation. To validate these labels, we randomly select $10\%$ of the data from both DEPTWEET and SemEval-2018 T3 and obtain annotations from native speakers with prior corpus annotation experience. The human annotators, all PhD students with domain expertise, followed the same detailed guideline used for automatic annotation.

\begin{figure}[h]
    \centering
    \begin{subfigure}{0.45\columnwidth}
        \centering
        \includegraphics[width=\linewidth]{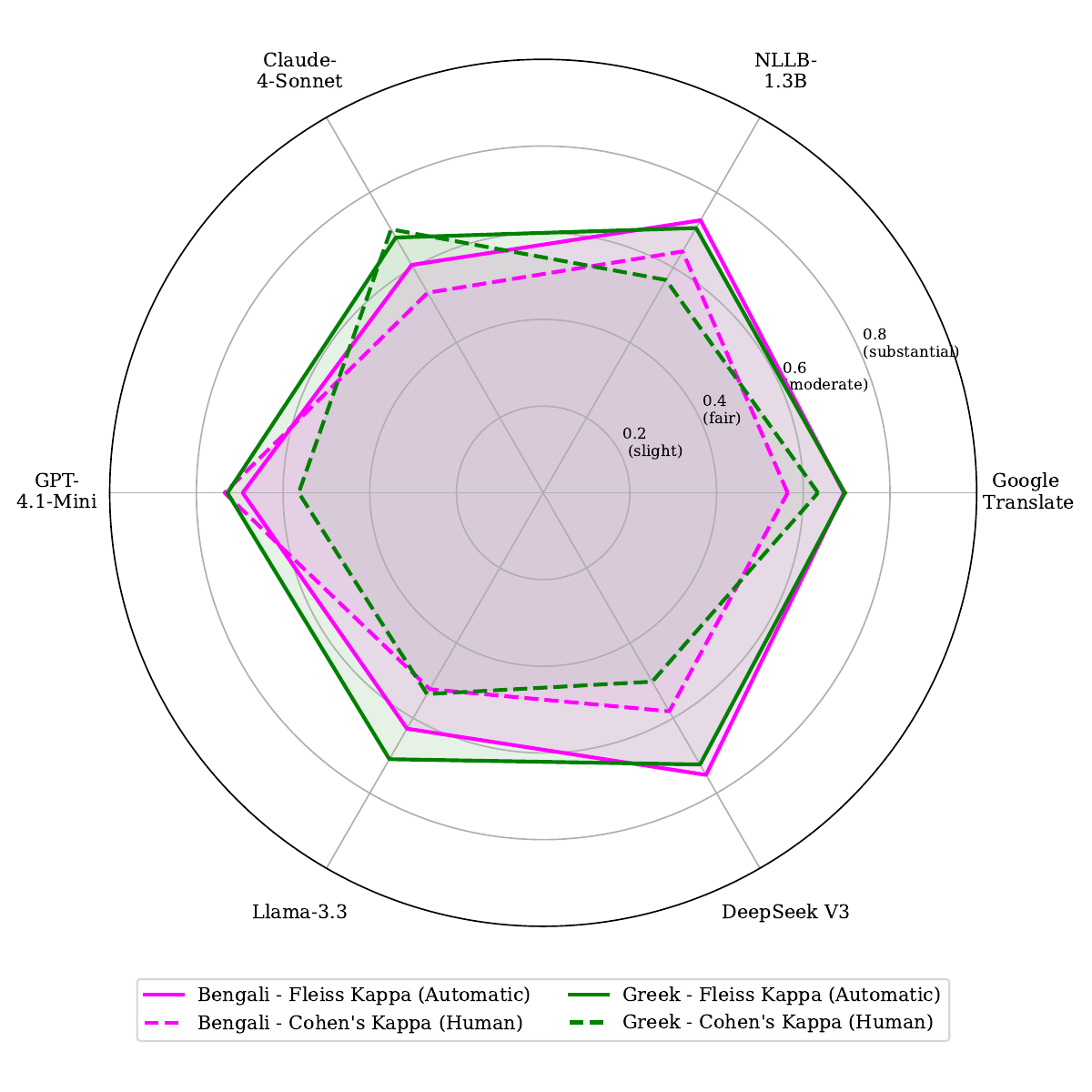}
        \caption{IAA Scores for DEPTWEET}
        \label{fig:sub1}
    \end{subfigure}
    \hfill
    \begin{subfigure}{0.45\columnwidth}
        \centering
        \includegraphics[width=\linewidth]{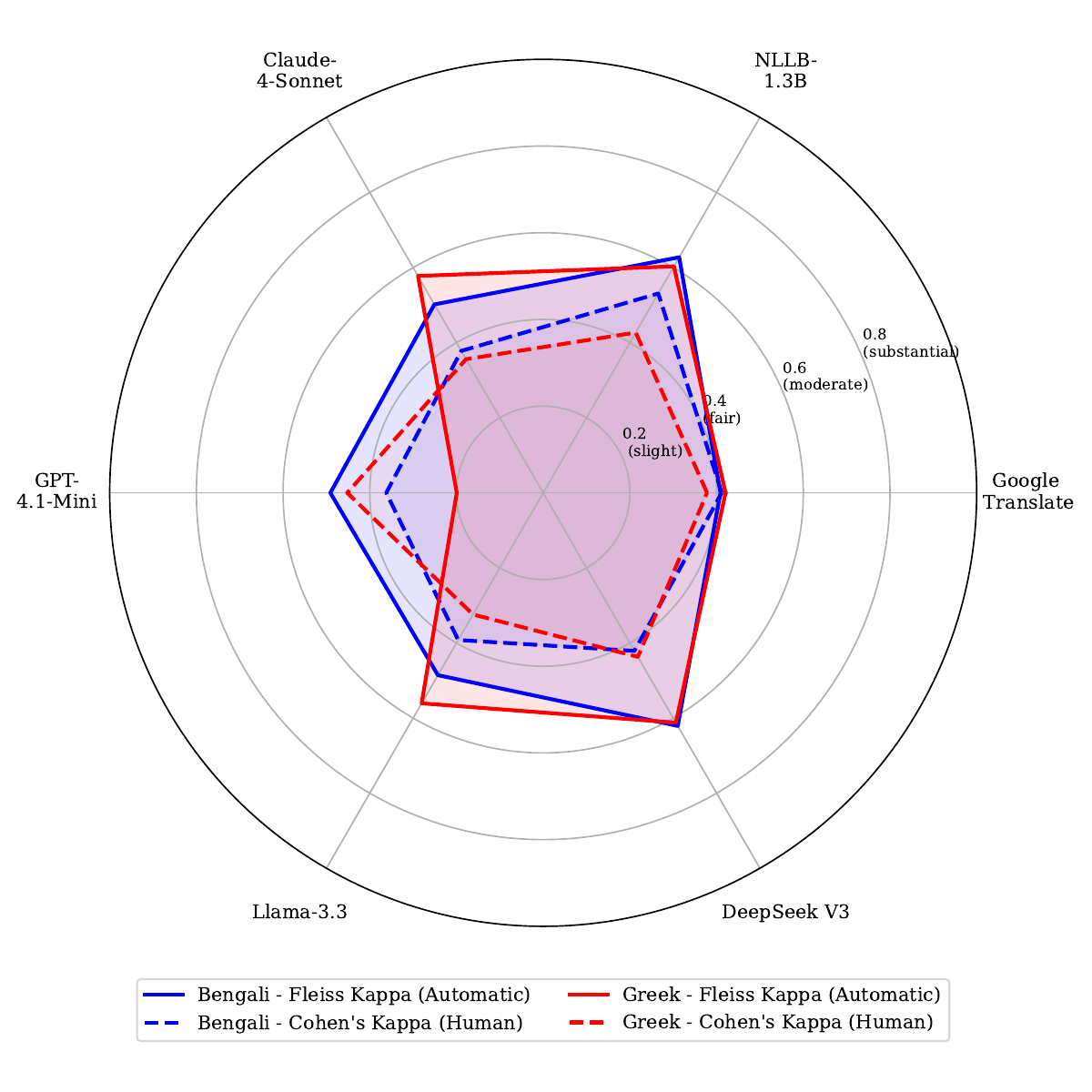}
        \caption{IAA Scores for SemEval-2018 T3}
        \label{fig:sub2}
    \end{subfigure}
    \caption{Agreement scores for automatic (solid) and human (dashed) annotations, ranging from fair to substantial agreement across both datasets.}
    \label{fig:agreement_scores}
\end{figure}

Inter-Annotator Agreement (IAA) is measured using Fleiss' Kappa among the three LLMs and Cohen's Kappa between the final automatic labels and human annotations. Figure \ref{fig:agreement_scores} presents the IAA scores. Automatic annotation achieves moderate to substantial agreement, while agreement between human and automatic labels ranges from fair to moderate for DEPTWEET. For SemEval-2018 T3, agreement remains fair to moderate between automatic and human annotations, reflecting the greater difficulty of this dataset.

\section{Cross-Cultural Translation: Beyond Label Preservation}

\subsection{Translation Refusal}

\begin{figure}[h]
    \centering    \includegraphics[width=\columnwidth]{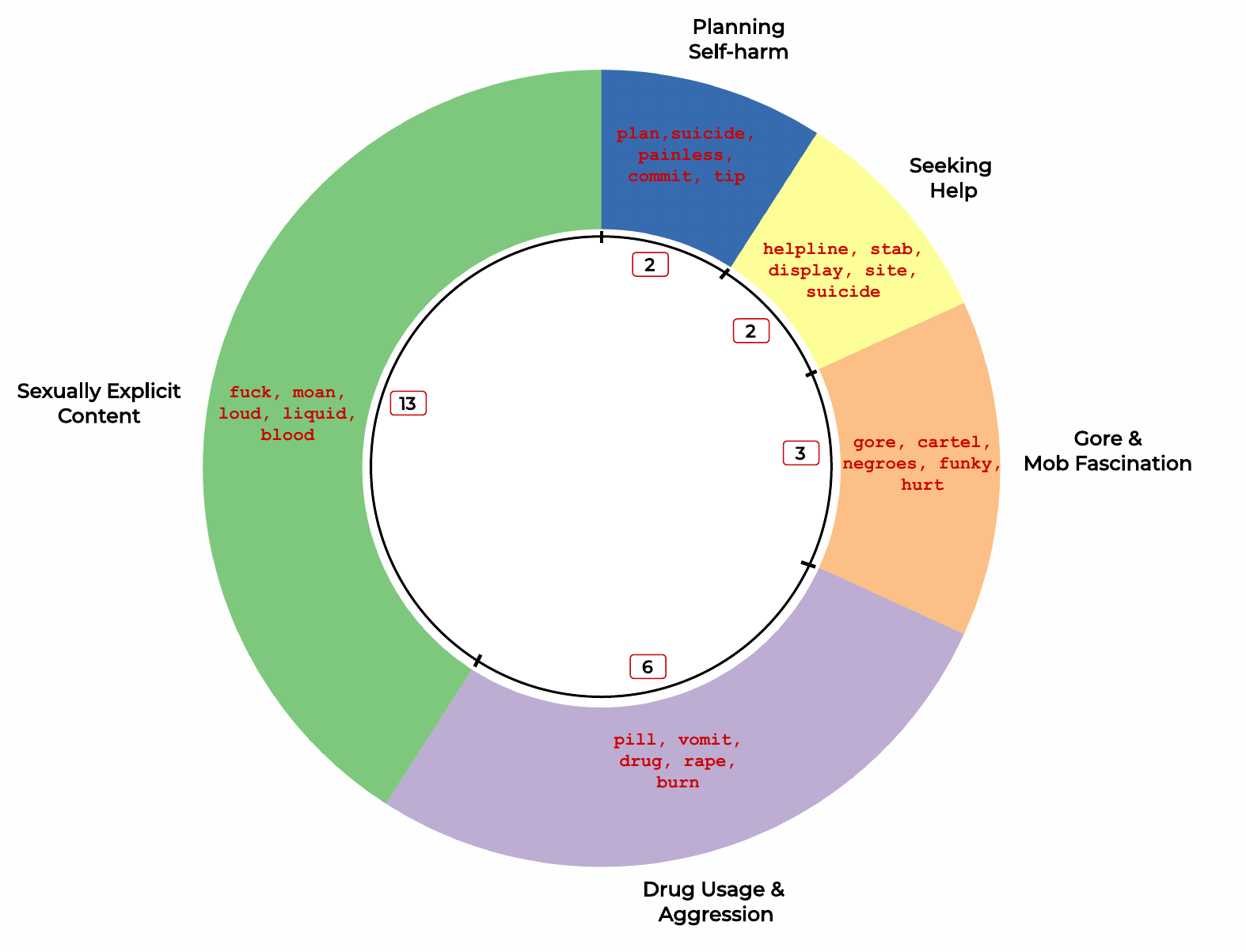}
    \caption{Topic distribution of refused samples showing five identified themes. Outer labels indicate topic categories, keywords inside represent the most characteristic terms, and center values show sample sizes (sentences per topic)}
    \label{fig:topic_refusal}
\end{figure}

We observe that Claude Sonnet 4 and DeepSeek V3 refuse to translate a subset of samples across both Bengali and Greek datasets. Notably, the same sentences are consistently rejected in both languages. We apply HDBSCAN clustering followed by BERTopic modeling to analyze these refused samples, as shown in Figure \ref{fig:topic_refusal}. The rejected content primarily involves sexually explicit material, slang and aggression, gore, and drug-related themes. Further inspection reveals that these sentences belong largely to the \textit{Severely Depressed} class ($\sim$60\%) in DEPTWEET and the \textit{Situational Irony} class ($\sim$37\%) in SemEval-2028 T3, both minority categories in their respective datasets. Since these samples carry rich but underrepresented information, their exclusion during cross-lingual translation risks losing valuable cultural and linguistic signals, a factor that requires careful consideration in cross-cultural MT.

\subsection{Cultural Misalignment}

We find a number of samples that contains cultural references in the specific culture, and gets completely lost when translated in Bengali or Greek. 

\noindent\includegraphics[width=\linewidth]{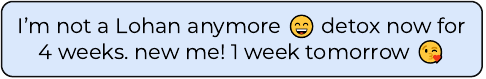} 
For instance, the phrase \textit{"I'm not a Lohan anymore"} in the sentence in the textbox (taken from SemEval-2018 T3) is a cultural reference to actress Lindsay Lohan and her public struggles. A Bengali or Greek reader unfamiliar with this context would struggle to interpret the underlying meaning. The translation fails to convey the meaning of "I've left my troubled past behind." Even the cultural prompting misses this to align with the respective culture.



Another frequent source of cultural misalignment arises from the use of slang. Slang expressions common in Western contexts can become profane or overtly vulgar when translated into more conservative cultures. 

\noindent\includegraphics[width=\linewidth]{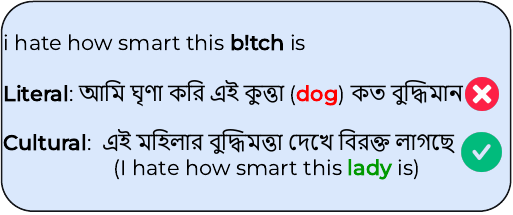}
For example, the above sample from the DEPTWEET dataset contains the term \textit{``b!tch''}, which in English is often used playfully, combining admiration with mild insult. However, in Bengali, the literal translation renders it as a highly offensive term. The culturally prompted version mitigates this by replacing the slang with a contextually appropriate feminine expression, aligning better with Bengali social norms. In Greek, the literal translation from NLLB-1.3B ``{\selectlanguage{greek}σκύλα}'' retains the insult but sounds harsher than intended, while GPT-4.1-Mini’s literal rendering ``{\selectlanguage{greek}πουτάνα}
'' amplifies vulgarity, losing the original nuance. Conversely, the cultural translation ``{\selectlanguage{greek}γυναίκα}'' (woman) softens the tone excessively, erasing both the playful and pejorative dimensions. These examples illustrate that slang requires careful handling in cross-cultural translation, a challenge that cultural prompting appears to manage more effectively. In summary, cultural adaptation during translation \citep{singh2024translating} has the potential to effectively resolve the issues of refusal and misalignment by maintaining both semantic fidelity and cultural appropriateness.

\section{Conclusion}

In this study, we investigate semantic label drift in cross-cultural translation. Focusing on two culturally sensitive domains- mental health and irony, we demonstrate that semantic labels often change after translation, both in traditional statistical machine translation systems (e.g., Google Translate) and in modern LLM–based translators. Beyond label drift, we identify a separate, mutually exclusive phenomenon: culturally inappropriate translations due to misalignment between the source and target language cultures. Our findings highlight the need to revalidate labels and apply cultural adaptation strategies before reusing translated datasets in cross-cultural NLP research.

\section*{Limitation}

This study focuses on two culturally sensitive domains: mental health and irony, to examine semantic label drift after translation. While we acknowledge that other domains such as sarcasm \citep{blasko2021saying} and humour \citep{jiang2019cultural} also exhibit strong cultural dependencies, their inclusion was beyond the scope of this work due to time and resource constraints. We select mental health and irony specifically because of the availability and granularity of existing annotated datasets in these areas.

Similarly, our cross-cultural experiments are conducted using Bengali and Greek as representative target languages. Although incorporating additional cultures would provide a broader perspective, our objective in addressing RQ3- \textit{examining whether cultural similarity between source and target languages mitigates label drift}, require a focused, controlled comparison. Expanding to more cultures could have diluted this comparative analysis. Moreover, access to qualified native annotators for verifying machine-translated labels is a practical constraint; Bengali and Greek offered the most feasible and representative choices under these conditions.

\bibliography{custom}

\end{document}